**Inferring Global Dynamics of a Black-Box System Using Machine Learning**

Hong Zhao
Department of Physics, Xiamen University,
Xiamen 361005, China
E-mail: zhaoh@xmu.edu.cn

We present that, instead of establishing the equations of motion, one can model-freely reveal the dynamical properties of a black-box system using a learning machine. Trained only by a segment of time series of a state variable recorded at present parameters values, the dynamics of the learning machine at different training stages can be mapped to the dynamics of the target system along a particular path in its parameter space, following an appropriate training strategy that monotonously decreases the cost function. This path is important, because along that, the primary dynamical properties of the target system will subsequently emerge, in the simple-to-complex order, matching closely the evolution law of certain self-evolved systems in nature. Why such a path can be reproduced is attributed to our training strategy. This particular function of the learning machine opens up a novel way to probe the global dynamical properties of a black-box system without artificially establish the equations of motion, and as such it might have countless applications. As an example, this method is applied to infer what dynamical stages a variable star has experienced and how it will evolve in future, by using the light curve observed presently.

## I. Introduction

Dynamical systems are the basic objects of scientific research. They have broad applications in a variety of disciplines ranging from physics, biology, chemistry, engineering, economics, medicine, and beyond. The typical approach for understanding a dynamical system is to establish its equation of motion, which can be described by $x(t) = f^t(x(0), \Lambda)$, where $x \in R^d$ is a $d$-dimensional vector variable, $t$ is the time, $f^t$ is the evolution operator, and $\Lambda$ is the parameter set. By investigating the entire parameter space one can grasp the global properties of the system. The inverse-problem approach, one of the most important mathematical tools in science, is aimed to fix the equation of motion by finding its parameters from a data set of observations [1-6]. However, to accurately model a real-world system is an extremely difficult problem – it needs accurate prior knowledge about the system.

Without the model, for a particular variable of the system, $x(t)$, if it is measured over a period of time, there are some model-free approaches that can be applied to predict the future evolution of this variable, such as the time series reconstruction method [7-8], empiricism-based method [9], deep belief network [10], long short-term memory network [11], dynamics reproduce method [12], and reservoir computing [13-18], etc. This kind of prediction is done, however, for variable states under a *fixed* system parameter set $\Lambda$. A more challenging task is to probe the *global* dynamical behavior of a wider parameter space of the system *behind* the time series in a model-free way under the same premise, i.e., to infer the dynamical properties that the system may exhibit in its parameter space, by using only time series measured at the *present* parameter set $\Lambda$.

Here we report that by employing the learning machine, this task can be fulfilled. We find that under an appropriate training strategy, the dynamics of the learning machine observed at different training stages may reproduce the global dynamics of the system behind the training data (we call the target system hereafter) along a special path in its parameter space. This path is particularly important since the global dynamical behavior of the target system, from simple to complex, may be qualitatively reproduced along that. The Lorentz system, one of the best-known models in nonlinear dynamics study, is adopted to demonstrate how our procedure works. The preliminary idea has been proposed by us in Ref. [12]. To the best of our knowledge, this approach has not been explored in the existing literature.

The mechanism if also revealed in this paper. We show that our particular training strategy could collapse the parameter space rapidly onto a subspace in which the learning machine appears to be equivalent to the target system. The parameter region reached in this early stage usually corresponds to simplest dynamics. With further training, the learning machine evolves to the target system with the present parameters, and may extend to parameter regions beyond the present parameter set, where more complex dynamical behaviors are expected. In this way, the learning machine reveals the global dynamics of the target system from simple to complex.

The simple-to-complex path is inherently consistent with the usual evolution law of natural systems. Therefore, taking the current stage as a reference point, the learning machine can not only reproduce the evolution history of a target system, but also predict how it will evolve in the future. For example, observations indicate that more than 30% of stars exhibit luminosity changes; variable stars are among them and have been used as the standard candles to measure the cosmic distance, but their evolutionary paths have not been fully understood yet. The time scale of human observation records is extremely short compared with the cosmological time scale, and thus it is impossible to explore the evolution process of the specific variable stars by direct observation. In the duration of records, the stars correspond to a fixed-parameter system. In the past decade, cosmological observations, particularly those from the Kepler space telescope, have yielded high-precision light curves of variable stars [19-22]. In such a background, we show that the learning machine provides a tool to infer the evolutionary path and to improve the classification of variable stars.

The paper is organized as follows. Section II introduces the learning machine we applied and explains how to train it. We emphasize that all parameters in our model are changeable, and we apply a Monte Carlo algorithm to adjust them. As such, we can decrease the cost function monotonically and thus enable the learning machine emerging different dynamics of the target system in different training stages. Section III validates the above picture using the paradigmatic Lorenz system. Another dynamical system with simple dynamics is also studied to emphasize that the dynamics emerged from the learning machine has intrinsic correlation to the behind system.

Section IV explains the mechanism why the learning machine can reveal the dynamics of the target system from simple to complex. The application to variable stars is presented in Sec. V. The last section discusses certain open problems.

## II. Learning machine and training strategies

The dynamics of our self-evolution learning machine can be constructed by various dynamical equations. In this paper two of them are used. One is a three-layer time-delayed map

$$\Phi(n+1) = \Phi(n) + \tau \sum_{i=1}^{N} u_i f\left(\beta_i(\sum_{j=1}^{M-1} v_{ij}\Phi(n-j) - b_i)\right), \quad (1)$$

where $M$ and $N$ are the number of input and hidden layer neurons, respectively, $\tau$ is the time interval between two sequential records of the target system's time series, and $f$ is the neuron transfer function. Besides $M$ and $N$, the learning machine contains four types of parameters, each is bounded in an interval, i.e., $|u_i| \leq c_u, |\beta_i| \leq c_\beta, |v_{ij}| \leq c_v, |b_i| \leq c_b$. We call $c_u, c_\beta, c_v, c_b$ training control parameters. The parameters are randomly initialized in their value ranges. Let $x(j), j = 1, \ldots, P$ be a segment of time series of variable $x$ of the target system at its present parameter set. We use it to construct $P$-$M$ training samples, with $\{\Phi(n-j) = x(n-j), j = M-1, \ldots, 0\}$ being the input of the $n$th sample and $x(n+1)$ the expected output. We define the cost function as

$$\lambda = \left(\frac{1}{P-M}\sum_{n=M}^{P-M}(x(n+1) - \Phi(n+1))^2\right)^{1/2}. \quad (2)$$

This learning machine is for reproducing the dynamics using one dimensional time series. Another one is a three-layer map

$$\boldsymbol{\Phi}(n+1) = \boldsymbol{\Phi}(n) + \tau \sum_{i=1}^{N} u_i f(\beta_i(\hat{\boldsymbol{v}}\boldsymbol{\Phi}(n) - b_i)), \quad (3)$$

where $\boldsymbol{\Phi}(n)$ is an $M$-dimensional vector. This learning machine is for reproducing the dynamics when all variables of the target system are available. The training samples can be constructed by $\{\text{input: } \boldsymbol{\Phi}(n) = \boldsymbol{x}(n), \text{output: } \boldsymbol{x}(n+1)\}$, and the cost is defined by

$$\lambda = \left(\frac{1}{P}\sum_{n=1}^{P}\sum_{j=1}^{M}\left(x_j(n+1) - \Phi_j(n+1)\right)^2\right)^{1/2}. \quad (4)$$

We apply a simple Monte Carlo algorithm to train the learning machine: Randomly mutating a parameter in its value range and accepting this variation if it does not increase $\lambda$. Each adaptation renews only $O(P)$ multiply-add operations and the adaptation accepted is optimum for the whole training set in the statistical sense. It does not need to evolve the entire network which needs about $O(NMP+NP)$ multiply-add operations [23,24]. As such, this algorithm is practical for applications except for those asking for immediate learning. Another advantage of using the Monte Carlo algorithm is that one can employ a variety of neuron transfer functions. In this paper, except for specially declared, we apply $f(h) = \exp(-h^2)$ to be the neuron transfer function.

With a proper set of the control parameters $c_u, c_\beta, c_v, c_b$ for fixed $M$ and $N$, we randomly initialize all parameters and perform the training using the Monte Carlo algorithm. The parameter $N$ controls the learning machine size. We demand that it should be sufficiently large, in which case it is insensitive to the result. Throughout this paper we take $N=3000$. After a fixed amount of Monte Carlo operations, the training is suspended with a new $\lambda$ reached. Then, after the learning machine has been self-evolved for a time long enough to go beyond the transient process, we record a sufficient long time series of $\Phi(n)$. This series represents the asymptotic dynamical behavior of the learning machine. One can represents the trajectory using delayed coordinates, $\Phi(n)$-$\Phi(n-Q)$, in two dimensional plane, where $Q$ is a proper integer. To represent the global dynamics as a function of the training process, we should apply the Poincare section representation. Supposing that the time series is recorded with sufficient short $\tau$, we can fit it to obtain a continuous variable $\Phi(t)$. A set of the delay coordinates, $\Phi(t_c-T)$, is obtained, where $t_c$ is the time that $\Phi(t)$ crosses the section of $\Phi(t)=c$ from above, and $T$ is the delayed time. In this way, we obtain the bifurcation diagram $\Phi(t_c-T)$- $1/\lambda$ of the learning machine along the axis of $1/\lambda$, which represents the global dynamics of the learning machine in different training stage. Our key finding is, as illustrated by instances in the next section, that the dynamics of the learning machine can be mapped to the dynamics of the target system along a particular path in its parameter space, and thus explores the global dynamical behavior of the target system around the present parameter set.

### III. Illustration examples
#### 3.1 The Lorenz system

The Lorentz system is given by $dx/dt = -\sigma(x-y), dy/dt = -xz + Rx - y$, and $dz/dt = xy - Bz$. Figure 1(a) shows the time series of $x$ variable at parameter set ($\sigma$, $R$, $B$) = (10, 28, 0.555). One can see that it represents a periodic trajectory. We apply a segment of time series of length $t=30$ recorded with the interval of $\tau=0.01$ to train the learning machine with $M=60$.

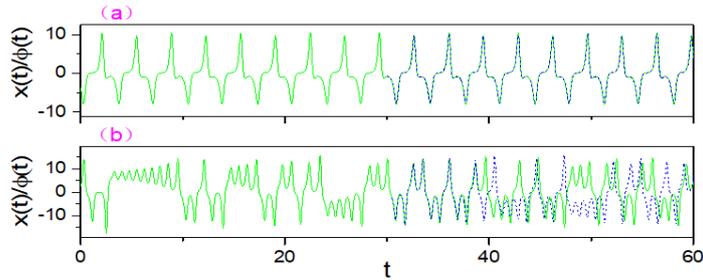

**Fig. 1.** Solid lines are time series of $x$ variable made at (a) $B=0.555$ and (b) $B=2$ respectively. The segment of time series of $t\leqslant 30$ are used for training. Dotted lines are predicted results.

Figure 2(a) shows the bifurcation diagram of the learning machine with other training control parameters fixed at $c_\beta = 0.5, c_u = 0.2, c_v = 2, c_b = 15$. The vertical axis represents $\Phi(t_c-0.01)$, which is the delayed coordinate of the section point $t_c$ that the $\Phi$ variable crosses the line of $\Phi=5$ from the above. We see that the learning machine shows a typical bifurcation diagram of nonlinear chaotic system, i.e., it develops to chaos through period-doubling bifurcation road. At the position marked by the digit 2 the learning machine can perfectly reproduce the training time series, i.e., by in putting a segment of $x$ variable with length $t=0.5$ ( $\tau M=0.5$) to start up the learning machine, the time series $\Phi(t)$ can then mimic the subsequent evolution of $x(t)$, as Fig. 1(a) shows. Therefore, the learning machine at this stage reproduces the current target system and predicts the variable evolution with high accuracy. With this function, the learning machine can predict the variable evolution of the present target system.

The question of higher interest is whether this bifurcation diagram is related to the dynamics of the Lorenz system, and thus can reveal the global dynamical behavior of this system in parameter space different from the present parameter set. Figure 2(b) shows a the bifurcation diagram of the Lorenz system along the parameter $B$ with other two parameters fixed at $(\sigma, R) = (10, 28)$. The vertical axis represents the delayed coordinate $x(t_c-0.01)$. Here $t_c$ is the time that the $x$ variable crosses the line of $x=5$ from the above. The training sample in Fig. 1(a) is obtained at the position marked by digit 2 in Fig. 2(b). It is a period-2 limit cycle, see Fig. 2(c). We find that the bifurcation diagram of the learning machine is highly similar to that of the Lorenz system in the interval of $B\in(0.45,B_c)$, even the envelop curve structure. Further convincing evidence is provided in Fig. 2(c), which shows that the trajectories along the two bifurcation diagrams have identical topology. We can therefore draw the conclusion that the learning machine proves to be equivalent to the Lorenz system in a region around the present parameter set where the training data is collected.

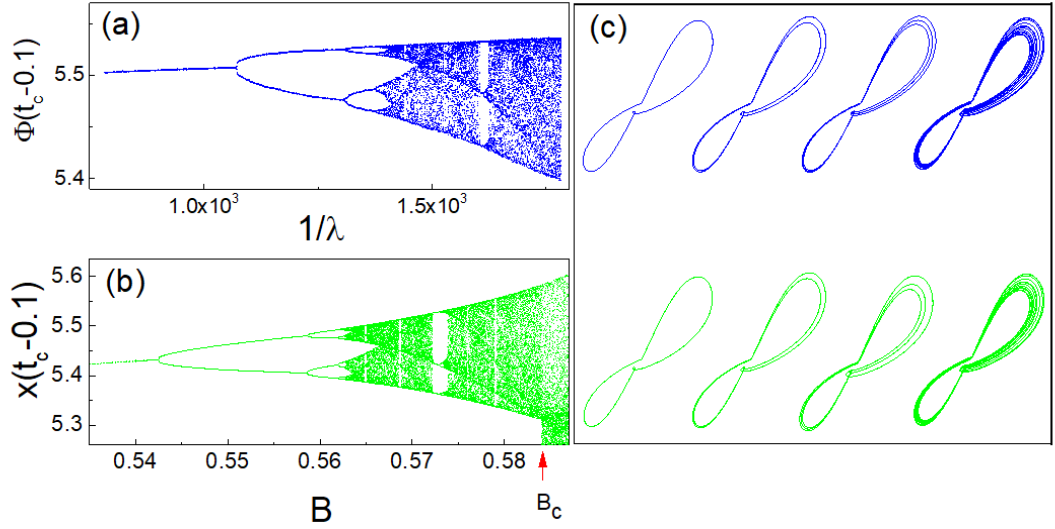

**Fig. 2.** Bifurcation diagrams of the learning machine (a) and the Lorenz system (b). (c) shows the corresponding trajectories along the two bifurcation diagrams. They are represented by $x(t)$-$x(t$-$0.1)$ with scales of $x(t)\in(-15,15)$.

We have checked that applying data samples from any position of $B\in(0.45,B_c)$ can reproduce the global dynamics in this region. Figure 3 (a) and 3(b) show the bifurcation diagrams of the learning machine trained by time series of $x$ variable measured at $B=0.5725$ and $B=0.58$, respectively. At $B=0.5725$ the Lorenz system shows a period-3 limit circle embedded in the chaotic region. The delayed-coordinate trajectory is shown in Fig. 2(c) and marked by digit 3. At $B=0.58$ the Lorenz system shows the chaotic motion, whose delayed-coordinate trajectory is marked by digit 4 in Fig. 2(c). We see that in both cases the global dynamics of the target system in the region of $B<B_c$ is reproduced. We have checked that no matter how close the sampling position is to $B_c$, the resulted bifurcation diagram does not cross the point of $B=B_c$. The failures past $B_c$ are manifested as a quick divergence or a sudden change in the amplitude of the output (the output no longer maintains a clear similarity with that of the target system). Note that $B=B_c$ is a crisis point, at which the two coexisting-attractor branches collide (see Fig. 4) and thus the amplitude of the trajectories shows a sudden burst. Therefore, the reproduce ability of the learning machine is limited by intrinsic properties of the target system, as the crisis does that can abruptly increase the complexity.

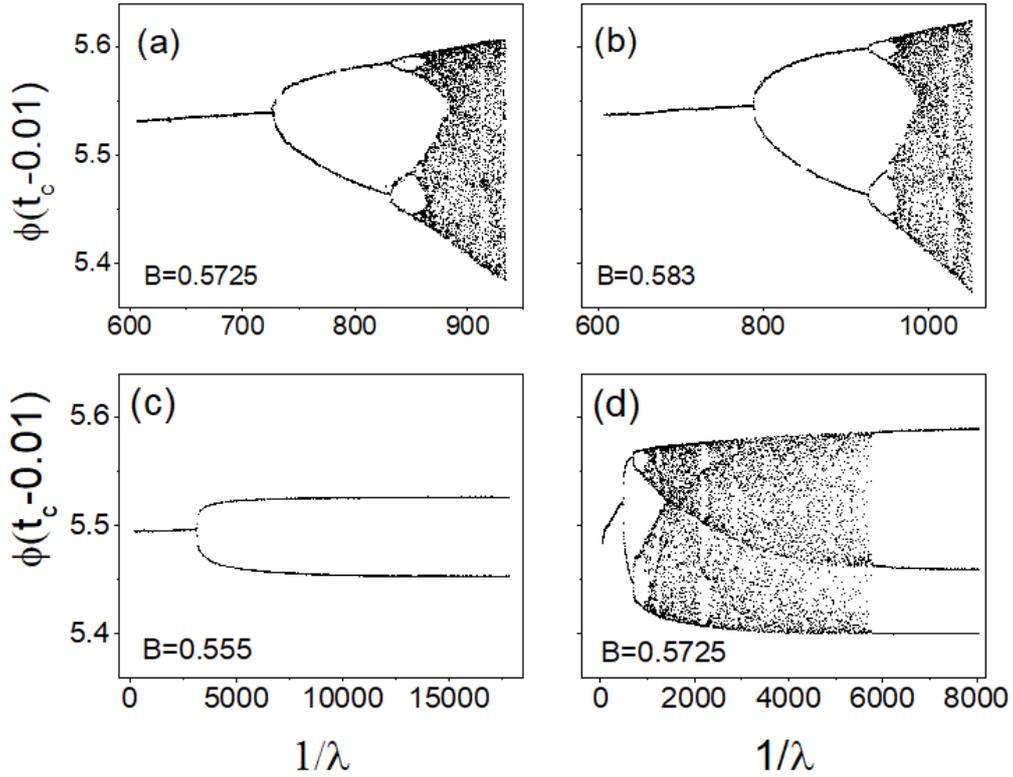

Fig. 3. Bifurcation diagrams of learning machines for training samples made at B=0.5725 (a), B=0.583 (b), B=0.555 (c), and B=0.5725 (d), respectively. Parameters used in (a) and (b) are $c_\beta = 0.45$ and $c_\beta = 0.4$ respectively and others are fixed at $c_u = 0.2, c_v = 2, c_b = 15$. Parameters used in (c) and (d) are $c_\beta = 0.4$ and $c_\beta = 0.3$ respectively and others are fixed at $c_u = 1, c_v = 1, c_b = 20$.

The region of the parameters that can be revealed by the learning machine depends also on the choice of control parameters. There are various combinations of training control parameters that can lead to the fully developed bifurcation diagram as Fig. 2(a). Meanwhile, many other combinations may only partially reproduce the bifurcation diagram, commonly encountered scenario is that the part from the period-one limit cycle to the present state is reproduced, as Fig. 3 (c) and 3(d) show.

The dynamics in the region of $B \in (0.45, B_c)$ is typical in nonlinear chaotic systems. It represents the simplest road that a nonlinear system develops to chaos. The global dynamics of the Lorenz system throughout the parameter space is very complicated. Fig. 4(a) shows the bifurcation diagram of the Lorenz system in the region of $B \in (0.35, 2.2)$ with another two parameters fixed at $(\sigma, R) = (10, 28)$. We see that there are many stages with coexisting attractor branches. Figure 1(b) indeed shows a segment of time series of $x$ variable at $B=2$, which represents the full developed chaotic motion. Figure 2(b) is indeed an enlargement of the upper branch in the interval of $B \in (0.45, 0.587)$ in Fig. 4(a).

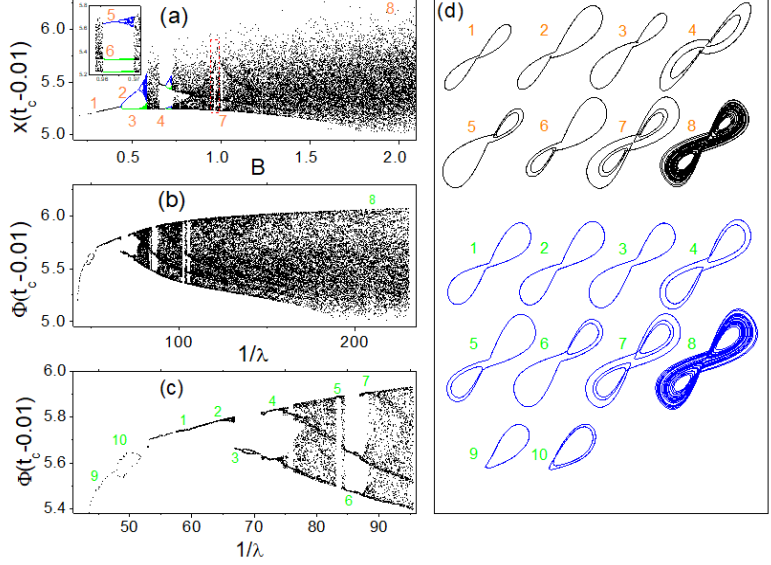

**Fig. 4.** (a) The bifurcation diagram of the Lorenz system along parameter $B$. The pieces colored in blue and green are coexisting attractor branches at the same parameter region. The inset is a zoom-in of the area marked by the red rectangle. (b) The bifurcation diagram of the learning machines trained by the sample shown in Fig. 1(b). (c) zooms in the beginning part of (b). (d) shows the trajectory samples from the Lorenz system and the learning machine, respectively. The corresponding trajectories are marked by the same digit. The coordinate scale for each plot is same as in Fig. 2(c).

We then investigate to what extent the learning machine can infer the global dynamics of the target system in such a wide parameter region. For this purpose, we apply the sample time series of length $t=30$ in Fig. 1(b) to train the learning machine with $M=60$. With the training control parameters $c_\beta = 0.2, c_u = 0.2, c_v = 2, c_b = 15$, the resulted bifurcation diagram of the learning machine is shown in Fig. 4(b). Figure 4(c) zooms in the beginning part of the bifurcation diagram. We see that Fig. 4(b) and Fig. 4(c) have obvious differences with the Lorenz system. However, by examining the trajectories at different stages, we find that the primary limit cycles that have appeared in period windows along the bifurcation diagram of the Lorenz system are reproduced, see Fig. 4(d). These solutions are distributed in the region of $B<1$, far from $B = 2$, at which the training data is collected. An extra segment in the beginning of the bifurcation diagram shows up and two trajectories, marked by digit 9 and 10 on this segment, are also shown in Fig. 4(d). Interestingly, by carefully checking the Lorenz system, we do find that there are solutions of the Lorenz system around $B=0.05$.

On the other hand, the learning machine loses certain information of the target system. The primary limit cycles remain but, taking the two period-1 limit cycles, marked by 2 and 7, for example, the subsequent period-doubling cascade is lost. In addition, the coexisting-attractor branches that appear simultaneously in the bifurcation diagram of the target system may appear discontinuously along the axis of $1/\lambda$ (see Fig. 4(c)), implying that the symmetry of the target system is not precisely kept. In addition, from Fig. 2(c), we see that the amplitudes of the predicted trajectories are inconsistent with those of the corresponding trajectories of the Lorenz system, except that for the current parameter set. The reproduction of the global dynamics is thus only qualitative.

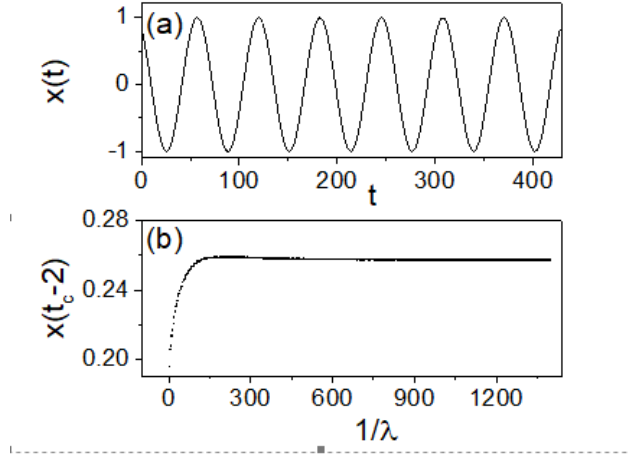

**Fig. 5.** The training sample of a simple nonlinear system (a) and the resulted bifurcation diagram of the learning machine (b).

### 3.2 The stable limit-cycle model

The model described by $dx/dt = x + y - x(x^2 + y^2)$ and $dy/dt = -x + y - y(x^2 + y^2)$ has a stable limit-cycle attractor, i.e., $x^2 + y^2 = 1$. As a two-dimensional ordinary differential equation system, it cannot have any chaotic motion. Therefore, its dynamics should be simple. Figure 5(a) shows a segment of time series of the *x* variable, which is a periodic oscillation and it is similar to Fig. 1(a). We apply this sample to train the learning machine with *M*=60 (the record interval is $\tau$ =0.01), the resulting bifurcation diagram of the learning machine is shown in Fig. 5(b). The training control parameters applied for this realization are $c_\beta = 0.2, c_u = 1, c_v = 1, c_b = 15$. We have checked that other combinations reach qualitatively the same results. It can be see that the bifurcation diagram converges to the present state quickly and keeps unchanged hereafter. This fact indicates that the learning machine does have an intrinsic connection to the target system, and the behavior emerged in different stages of the training process is closely related to the global dynamics of the target system in different parameter region.

## IV. The mechanism

We explain why applying only the data obtained at the present state the learning machine can subsequently emerge the primary dynamics of the target system in the simple-to-complex order. To give a clear physical picture, we employ the learning machine (3) to reveal the mechanism. Figure 6(a) and 6(b) show the bifurcation diagrams of the learning machine trained by using the records of full variables at ($\sigma$, $R$, $B$) = (10, 28, 0.5725), which is a period-3 limit cycle embedded in the chaotic region. The record interval is also $\tau$=0.01 and the length of the records is $t$=10. The neuron transfer function adopted in Fig. 6(a) is $f(h) = h^3$ and in Fig. 6(b) is $f(h) = \exp(-h^2)$ respectively. The training controlling parameters for the former are $c_\beta = 0.2, c_u = 0.2, c_v = 1, c_b = 15$ and for the latter are $c_\beta = 0.15, c_u = 0.2, c_v = 1, c_b = 5$. We see that in the first plot the bifurcation diagram of Fig. 2(b) ends at the period-3 window, while in the second plot the reproduce goes beyond the period-3 window.

The learning machine with $f(h) = h^3$ can be written explicitly, as $x(n + 1) = x(n) + \tau \sum_{i=1}^{N} u_i \left(\beta_i(v_{i1}x(n) + v_{i2}y(n) + v_{i3}z(n) - b_i\right)^3$ for the first equation of motion for example. The summation part can be expanded into 20 items, with 10 cubic, 6 quadratic, 3 linear ones. In Fig. 6(c) we plot the 20 coefficients of the 20 items of the summation part in the training process for this equation. Note that by Euler integration the first equation of the Lorenz system appears as $x(n + 1) = x(n) + \tau(-\sigma x + \sigma y)$. By an overall view, we see that the learning machine collapses quickly to the Lorenz system, i.e., the coefficients of the $x$ and $y$ variables converge towards $\pm\sigma = \pm 10$ while other coefficients tend to zero. A close observation reveals that, as the inset in Fig. 6(c) shows, not all of them strictly vanish. These non-vanishing coefficients, even for those below 0.1%, cannot be ignored. Otherwise, the trajectories will behave substantially differently from the target system. These facts indicate that the collapsed learning machine, on one hand, should have qualitatively the same dynamical behavior with the target system, since their dominant terms are the same. On the other hand, what it reproduces is a qualitatively similar system, not an identical one.

Why the learning machine in the initial training stage converges rapidly to a low-period limit cycle, which is corresponding to a simple dynamical state of the target system, can be understood in general. Based on dynamical system theory, attractors of low-period limit cycles have smaller Lyapunov exponents than those of high-period limit cycles or chaotic attractors. Our training seeks a fast convergence of the cost function, which is actually measured by the convergence rate of nearby trajectories. In the initial stage of the training, the trajectory of the learning machine still keeps a certain distance from the training trajectory. In this case, the training sample provides qualitative constraints to the learning machine, and its fine structure has not yet played a role. In this stage, converging to the parameter region of low-period limit cycles can result in a fast decrease of the cost function. As the training progresses further, the finer structure begins to play a more and more important role, forcing the learning machine to evolve towards the present state that the training sample is made. As a result, the training process draws a path in the parameter space along which the system evolves from simple to complex.

The degree of equivalence determines the scope of reproduction. In Fig. 6(a), the reproduction ends at the corresponding state where the training data is collected. As we apply $f(h) = h^3$ to be the neuron transfer function, the equations of motion of the learning machine are polynomial functions, matching those of the target system. As a result, the cost function can be decreased to become close to zero ($\lambda\sim10^{-4}$, see Fig. 6(a)). Correspondingly, the difference between the predicted trajectory and the training time series is also reduced to the same order of magnitude, indicating that the learning machine has closely approach the target system with the present parameters values. As the cost function has close to zero, further decreasing it cannot significantly change the parameters of the learning machine and thus cannot lead to any further modification in the dynamics. Mapping the path to the parameter space of the target system, it is a path ends at the present state.

While applying $f(h) = \exp(-h^2)$ as the neuron transfer function, the cost function that can be reached is larger ($\lambda\sim10^{-2}$, see Fig. 6(b), the section points become denser at the right end, indicating that further decreasing the cost function becomes more and more difficult). In such a case, the learning machine cannot be an accurate approximation of the target system. Nevertheless, since having already approach to the target system (in the order of $\lambda\sim10^{-2}$), the path that it follows is corresponding to a different path in the parameter space of the target system which does not end at the point of present parameters values, and thus it may reveal more abundant information of the target system.

The results shown in Fig. 3 can be understood accordingly. Although only $f(h) = \exp(-h^2)$ is applied (Note that the learning machine there is different from that adopted in Fig. 6(b)), one can properly adjust the training control parameters to adjust the precision of the approximation. Relative low approximation [Fig. 3(a) and Fig. 3(b), manifested by the relative big value of cost function] can reproduce the bifurcation diagram in the entire interval of $B\in(0.45, B_c)$, while high precision approximation [Fig. 3(c) and Fig. 3(d), manifested by the relative small value of cost function] can only reproduce the region from the period-1 limit cycle to the present state.

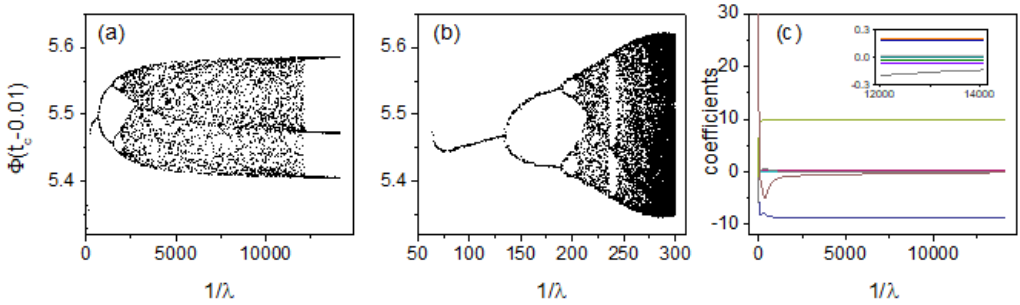

**Figure 6:** Reproduce using $f(h) = h^3$ (a) and using $f(h) = \exp(-h^2)$ (b). (c) shows the evolution of the 20 coefficients in the first equation of motion of the learning machine using the first neuron transfer function. The insert zooms in those coefficients close to zero in an interval.

## V. Application to variable stars

The Kepler space telescope provides the best light curves of variable stars so far [19-22]. There are 16 measurement modules in the database. Due to instrument adjustment and other reasons, there are big biases between different module data. In order to avoid the biases, we only use the data in one module as a training sample to avoid possible deception due to preprocessing. As such, the length of our training data is within 3 months. We downloaded the Kepler data from the Mikulski Archive for Space Telescopes [25].

Figure 7(a) shows the light curve samples for a so-called Blazhko-effect variable stars KPL7198959. This kind of stars are common among variable stars. The Blazhko effect refers to the phenomenon that the light curve has a periodic amplitude and/or phase modulation. The modulation period may be single-periodic, multi-periodic or irregular [26, 27]. Certain Blazhko stars pulsate intermittently with double period of the fundamental one [28, 29]. The KPL7198959 is such an example. From the sample light curve, one can see that the period-doubling characteristics emerges definitely around the left-hand side corner as well as an interval around t=30 day in Fig. 7(a). The emergence can repeatedly occur in the subsequent time. This phenomenon is different from the conventional period-doubling trajectories showing continuous doubled period.

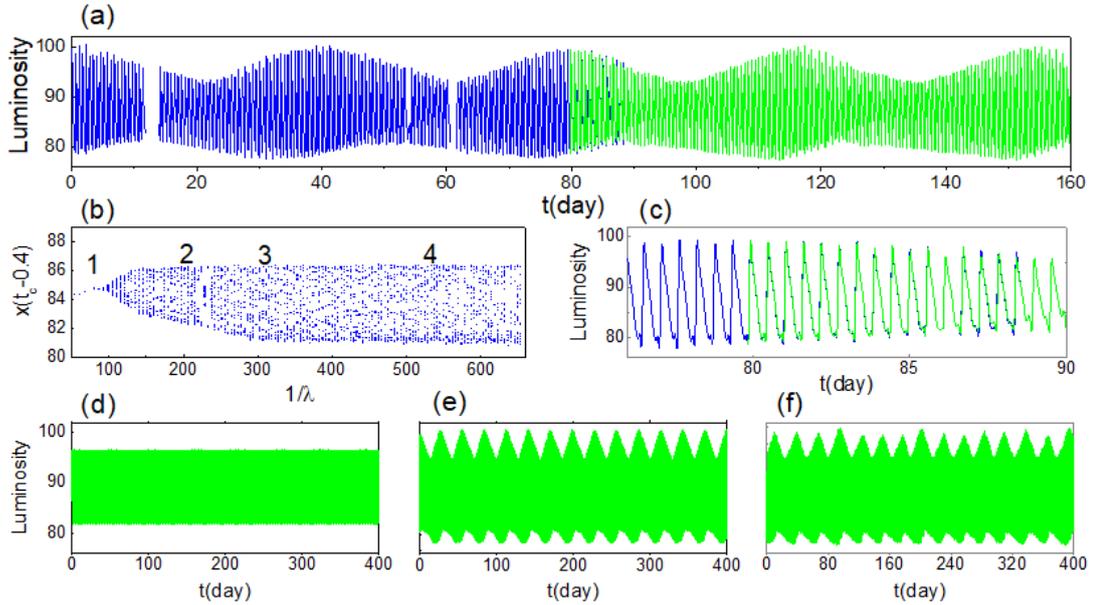

**Fig. 7.** (a) The solid blue curve is the observed light curve of KPL 7198959, and the dashed green one is predicted by the learning machine. (b) The bifurcation diagram of the learning machine caused by the observed light curve with $t \leq 80$. (c) Zooms in a segment around $t=80$ in Fig. 8(a). (d), (e) and (f) show the patterns of light curves of the learning machine at positions denoted by digits 1, 2, and 4, respectively.

Figure 7(b) shows the bifurcation diagram of the learning machine resulting from the training light curve of the first 80 days. Gaps represent missing raw data. The training control parameters are $c_\beta = 0.05, c_u = 1, c_v = 1, c_b = 50$, and $M = 50$. The amplitude of light curves are renormalized to the interval [-1,1] in training. Figure 7(a) also shows the predicted light curve of the learning machine with cost function value denoted by digit 3 in Fig. 7(b). Figure 7(c) zooms in a segment around t=80 day. We see that it can perfectly predict the future light curve for at least 8 days, afterwards the predicted curve keeps high similarity to the light curve of the star. The Blazhko effect remains, and particularly, the period-doubling characteristics emerges in an intermittent manner, consistent with the observed phenomenon.

Therefore, the learning machine with this cost function reproduces the present light-curve dynamics of the star. The bifurcation diagram before the digit 3 shows two stages. The start part represents a pure single-cycle pulsation without the Blazhko effect, as Fig. 7(d) shows. Then it evolves into a Blazhko-effect stage via a Hopf bifurcation. The Blazhko effect in this stage shows no period-doubling characteristics; it appears as a single-periodic modulation, as Fig. 7(e) shows. After the position of digit 3, the modulation mode changes eventually. The pulsation may develop into one of multi-cycle or even an irregularly modulated Blazhko-effect stage. Figure 7(f) shows an example of irregularly modulation. The period-doubling characteristics keeps to a certain extent in these stages. One cannot confirm these prediction by direct observation. However, from the Kepler database one can find that stars with all of these light curve characteristics exist. Therefore, we infer that the bifurcation diagram represents not only the 'past' but also the 'future' of the star. Meanwhile, we have checked the Blazhko-effect stars without period-doubling characteristic, such as KPL5559631. We find from the bifurcation diagram that the modulation period remains single periodic, though the amplitude changes significantly. As a result, we can infer that a Blazhko-effect star with period-doubling characteristic may experience the single-periodic modulation stage, while the conclusion that a Blazhko-effect star without period-doubling characteristic may evolve to the stage with period-doubling characteristic cannot be drawn.

   Recently, another interesting type of variable stars, so-called Golden stars are found [30, 31]. This kind of variable stars pulsates with two principal frequencies close to the golden ratio. The second pulsation frequency $f$ to the fundamental frequency $f_0$ is about $f_0/f \sim 0.618$ [31]. Figure 8(a) shows the light curve sample for such a star, KPL5520878. The bifurcation diagram resulting from this sample is shown in Fig. 8(b). The control parameters used here are $c_\beta = 0.2, c_u = 1, c_v = 1, c_b = 50$, and $M = 50$. The amplitude of light curves are renormalized to the interval [-1,1] in training. We see that this star has a simple evolution dynamics. The position marked by digit 2 represents the present stage, as the learning machine at this position can accurately predict its light curve for all the investigated time (see Fig. 8(a)), indicating that its pulsation is not chaotic. Meanwhile, the evolution has experienced only two stages, i.e., from a quasi-periodic stage at which the amplitude of the pulsation varies slightly to the next quasi-periodic stage at which it varies distinctly. We have checked that there is no Blazhko effect at both stages; the patterns marked by green digits 1, 2 and 3 look like the pattern marked by green digit 1 in Fig. 7(d). During the evolution process, the fundamental frequency remains approximately the same, and another so-called golden ratio second pulsation mode, i.e., the ratio of its frequency $f$ to the fundamental frequency $f_0$ is close to the golden ratio ($f_0/f \sim 0.63$), coexists as well in both stages, as Fig. 8(d)-8(f) show.

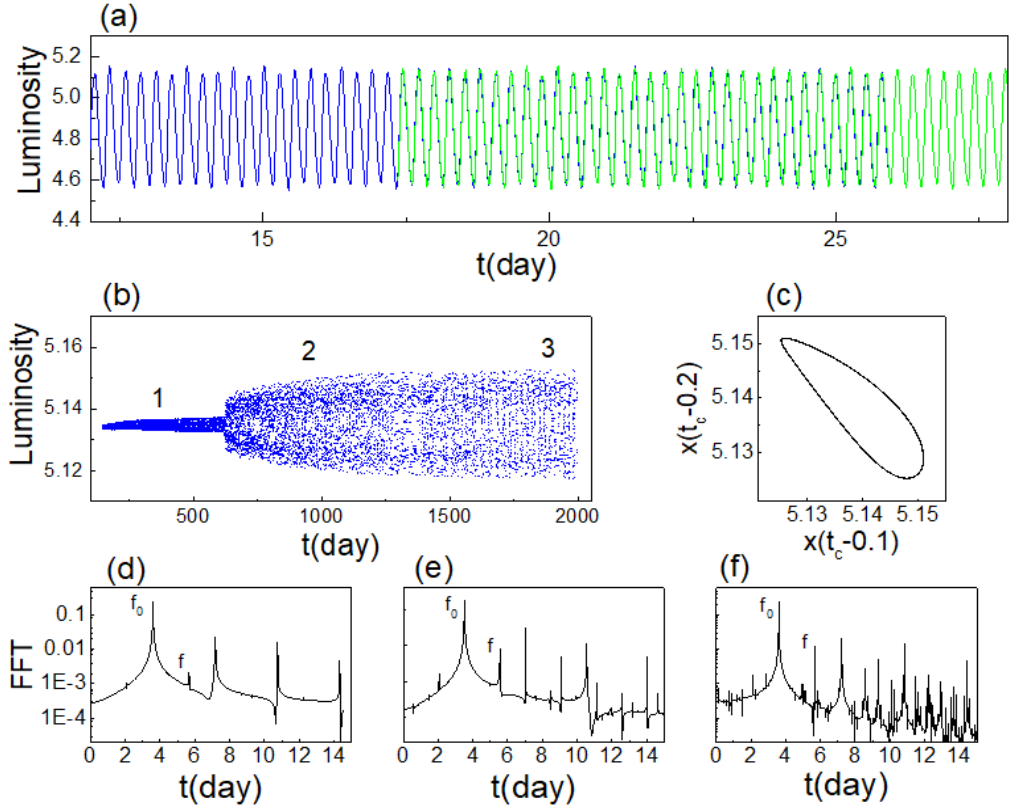

**Fig. 8.** (a) The solid blue curve is the observed light curve of KPL 5520878, and the dashed green one is predicted by the learning machine. (b) The bifurcation diagram of the learning machine caused by the observed light curve with t ≤ 17. (c) The section map of the predicted light curve. (d), (e) and (f) show the FFT of light curves of the learning machine at positions denoted by digits 1, 2, and 3, respectively.

The reproduced light curve can be used to clarify certain points. In Ref. [31] the star KPL5520878 is considered as an example of 'strange but non-chaotic motion'. The authors collected all of the data from 16 measurement modules in the Kepler database and found that the attractor shows a fractal geometry (strange) but the largest Lyapunov exponent is non-positive (non-chaotic). In their treatment, the row data has measurement error, and adjusting the biases of instrument in different modules may bring errors. A group of authors revisited the evidence of chaos in X-ray light curves of variable stars and find that the evidence of chaotic features may be due to the inaccurate data. To make a reliable conclusion one needs sufficiently long and stationary light curves [32]. They guess that this limitation may affect all similar investigations of light curves from other astrophysical sources.

The light curve of the learning machine avoids the limitation, as one can reproduce sufficiently long stationary measurement-error free time series. In Fig. 8(c), we plot a section map $x(t_c-0.1)$-- $x(t_c-0.2)$ of the predicted light curve of the learning machine at the position of digit 2. Here $t_c$ represents the time at which the light curve $x(t)$ crosses the section $x(t)=4.8$ from above. We see that the section is indeed a perfect limit cycle.

No fractal feature is found. Therefore, we believe the previous result may be due to the errors of data. We have checked some other Golden stars, such as KPL4064484, and have found similar phenomena.

## VI. Discussion

The new function of the learning machine introduced in this paper provides a model-free framework for inferring the global dynamics of black-box systems. This function is different from the traditional model-free methods of predicting variable evolutions. The latter are meant to predict the evolution of variables under chaotic conditions. Differently, our learning machine can emerge the primary dynamical properties of black-box systems, in the simple-to-complex order. We emphasize that the training sample is same as that used for the variable evolution prediction, i.e., we use only a segment of time serious recorded at present state of the black-box system. The present state represents the dynamics of the system at a point in the parameter space, while the revealed dynamics by the learning machine represents the primary dynamical properties overall at least a parameter region.

The mechanism we have introduced is the following. The time series of present state can impose restrictions on the global dynamics of the learning machine. With the monotonic decrease of the cost function, the parameter space collapses onto a subspace in which the learning machine appears to be approximately equivalent to the target system. The region initially reached usually has simple dynamics, such as limit cycles. This is due to the fact that this kind of attractors have the largest negative Lyapunov exponent, and thus induce a fast convergence of the cost function. In the subsequent evolution, the learning machine develops towards the parameter region corresponding to the present state, and thus the dynamics emerges in the simple-to-complex order. On the other hand, the equivalent of the learning machine to the target system is approximate. Mapping the evolution path of the learning machine into the parameter space of the target system, it is corresponding to a path that may end at the parameter point of the present state if the approximation degree is sufficiently high, or beyond the point of the present state if the degree is relatively low. In either cases, the emerged dynamical behavior qualitatively represents that of the target system. In this sense, the learning machine with a proper degree of equivalence may be more favorable since it allows one to probe the target system in a broader region. As an illustration we showed that a mere period-two limit cycle of the Lorenz system can be used to infer its period-doubling road to chaos.

The simple to complex path provides a way to inferring the evolution process of certain natural systems. Here we consider variable stars as an application, since their history cannot be verified by direct observation. The dynamics followed by the training process of the learning machine may provide important information to infer their evolutionary path, and to put restrictions to variable star theories, as well as to improve their classification.

One of our technique advantages is that we train the learning machine using the Monte Carlo scheme. This scheme can adjust all of the parameters of the learning machine and, particularly, can guarantee the monotonic decrease of the cost function.

As such, the learning machine reveals the dynamics of the target system from the simple to complex. Furthermore, this scheme avoids the restriction on the transfer functions. One can apply proper transfer functions to reach a better approximation to the target system. We would like to point out that, inferring the global dynamics of black-box systems may be done by other learning machines, if their training process are suitably designed.

Finally, we point out that this framework provides a different idea for the inverse problem approach. Note that the learning machine approach has been applied to many inverse problems [3-6]. Nevertheless, the motivation of those works is to determine the parameters of the model of the target system itself, following the standard approach of the inverse problem. However, inverse problems may have multiple solutions [1], which is almost inevitable in higher dimensional systems [2]. Consequently, different dynamical systems can produce the same output time series. In other words, there are different systems whose dynamics are equivalent to or approximately equivalent to a specific target system. Therefore, instead of finding the exact equations of motion of the target system, one can train a learning machine to approach one of such equivalent systems.

## ACKNOWLEDGMENTS

We thank Jiao Wang, Jie Yan, Lamberto Rondoni and Liang Huang for useful discussions. This work is supported by NSFC (Grant No. 11335006).